\documentclass[letterpaper]{article} % DO NOT CHANGE THIS
\usepackage{aaai23}  % DO NOT CHANGE THIS
\usepackage{times}  % DO NOT CHANGE THIS
\usepackage{helvet}  % DO NOT CHANGE THIS
\usepackage{courier}  % DO NOT CHANGE THIS
\usepackage[hyphens]{url}  % DO NOT CHANGE THIS
\usepackage{graphicx} % DO NOT CHANGE THIS
\usepackage{natbib}  % DO NOT CHANGE THIS AND DO NOT ADD ANY OPTIONS TO IT
\usepackage{caption} % DO NOT CHANGE THIS AND DO NOT ADD ANY OPTIONS TO IT
\usepackage{algorithm}
\usepackage{algorithmic}
\usepackage{multirow}
\usepackage{amsmath,amssymb,amsfonts,eucal}
\usepackage{relsize}
\usepackage{subfigure}
\usepackage{newfloat}
\usepackage{listings}
\DeclareCaptionStyle{ruled}{labelfont=normalfont,labelsep=colon,strut=off} % DO NOT CHANGE THIS
\floatstyle{ruled}
\newfloat{listing}{tb}{lst}{}
\floatname{listing}{Listing}

\title{Constructing Robust Emotional State-based Feature with a Novel Voting Scheme for Multi-modal Deception Detection in Videos}

\author{
Jun-Teng Yang,\textsuperscript{\rm 1}
Guei-Ming Liu,\textsuperscript{\rm 1}
Scott C.-H. Huang,\textsuperscript{\rm 2}
}
\affiliations{
\textsuperscript{\rm 1} Institute of Communications Engineering, National Tsing Hua University, Hsinchu 30013, Taiwan \\
\textsuperscript{\rm 2} Department of Electrical Engineering, National Tsing Hua University, Hsinchu 30013, Taiwan \\

\{junteng0211, geuimingliu\}@gapp.nthu.edu.tw, chhuang@ee.nthu.edu.tw

}
\usepackage{bibentry}
\begin{document}

\maketitle

\begin{abstract}
Deception detection is an important task that has been a hot research topic due to its potential applications. It can be applied in many areas, from national security (e.g., airport security, jurisprudence, and law enforcement) to real-life applications (e.g., business and computer vision). However, some critical problems still exist and are worth more investigation. One of the significant challenges in the deception detection tasks is the data scarcity problem. Until now, only one multi-modal benchmark open dataset for human deception detection has been released, which contains 121 video clips for deception detection (i.e., 61 for deceptive class and 60 for truthful class). Such an amount of data is hard to drive deep neural network-based methods. Hence, those existing models often suffer from overfitting problems and low generalization ability. Moreover, the ground truth data contains some unusable frames for many factors. However, most of the literature did not pay attention to these problems. Therefore, in this paper, we design a series of data preprocessing methods to deal with the aforementioned problem first. Then, we propose a multi-modal deception detection framework to construct our novel emotional state-based feature and use the open toolkit openSMILE to extract the features from the audio modality. We also design a voting scheme to combine the emotional states information obtained from visual and audio modalities. Finally, we can determine the novel emotion state transformation (EST) feature with our self-designed algorithms. In the experiment, we conduct the critical analysis and comparison of the proposed methods with the state-of-the-art multi-modal deception detection methods. The experimental results show that the overall performance of multi-modal deception detection has a significant improvement in the accuracy from 87.77\% to 92.78\% and the ROC-AUC from 0.9221 to 0.9265.
\end{abstract}

% -----Framework Figure---------
\begin{figure*}[htbp]
    \centering
    \includegraphics[width=1\textwidth]{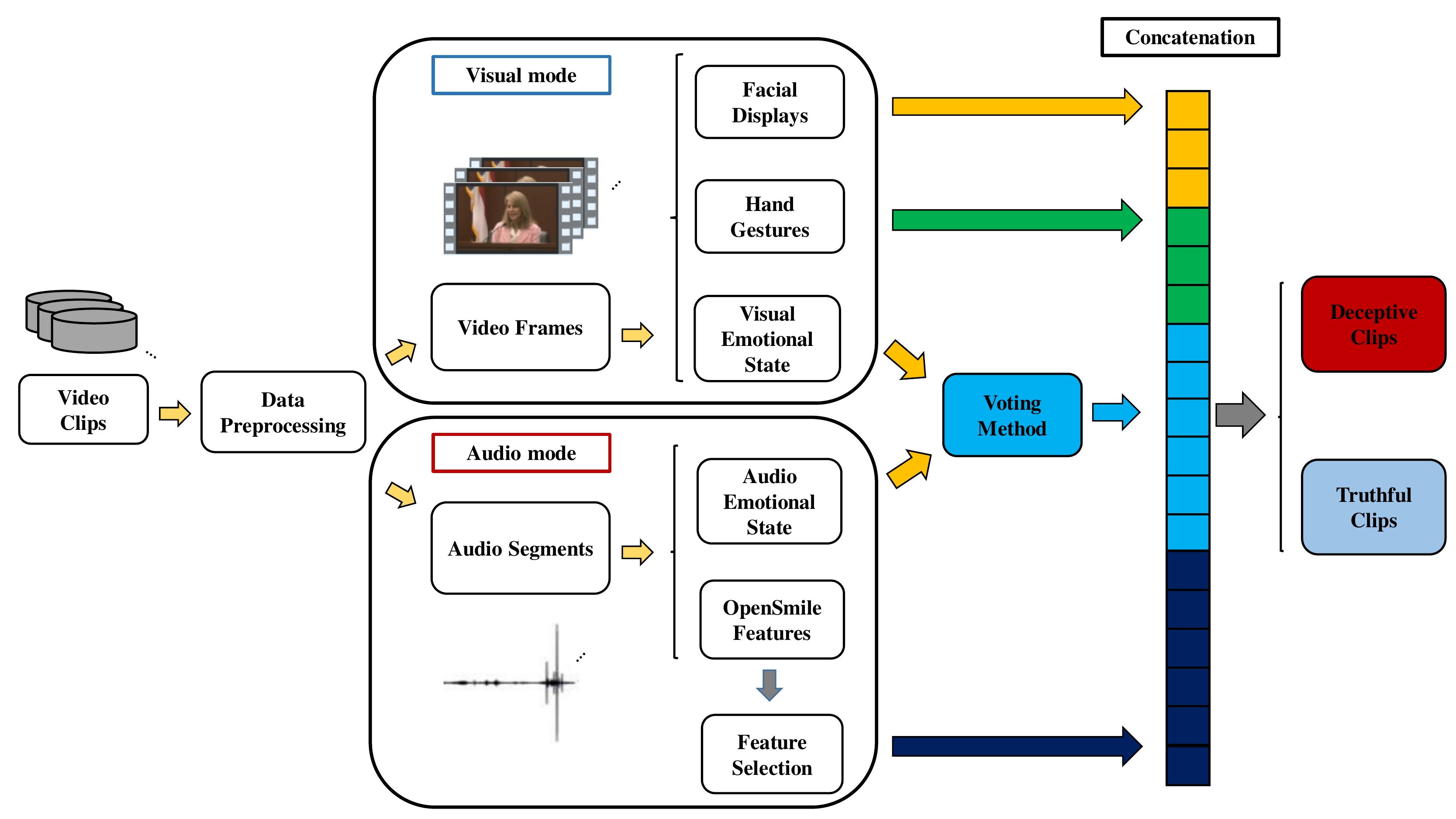}
    \caption{Our proposed multi-modal deception detection framework. We consider the features from visual and audio modalities to enhance the diversity of the training features.}
\label{fig.framework}
\end{figure*}

% ----- Introduction -----
\section{Introduction}
\label{sec:intro}
Human deception frequently occurs in our daily life. Although some deceits are innocuous, others may cause severe consequences and may become a fatal threat to enterprises, social security, and even countries. For example, deceiving in a court may influence the judiciary's judgment and cause a guilty accused to be acquitted~\cite{perez2015deception, mihalcea2012towards}. Therefore, accurate methods for detecting deceits are essential to personal and public safety in nowadays society. However, deception detection theory has attracted many researchers from different fields, such as jurisprudence, law enforcement, business, and national security, to build up human deception detection-related applications in the past decades~\cite{an2015literature}. Some physiological-based methods such as Polygraph and functional Magnetic Resonance Imaging (fMRI) have been designed for it, but the results are not always correlated with deceits~\cite{farah2014functional}. Moreover, since the results rely on the judgment of some professionals, they are often biased by individual differences. Hence, some outcomes in the court trials will be misjudged if the bands receive biased reports, and people who are not guilty will be put in prison while the guilty ones will be released. 

Recently, identifying behavioral clues for deception detection has become a hot issue for researchers in psychology, security, criminology, and computer vision fields \cite{michael2010motion, jaiswal2016truth}. Those behavioral clues can be categorized into verbal features (e.g., phonetic tones, self-reference, and cognitive words) and non-verbal features (e.g., facial expressions and body movements). Especially video is a friendly data type for deception detection tasks due to its naturally multi-modal property (e.g., visual, audio, and textual modalities). In addition, features extracted from both visual and audio modalities are rich in information related to human deception \cite{al2000detection, bond1990lie}. However, there are some challenges in both the feature extraction and the computational efficiency stages for deception detection in videos since video data contains complicated information in spatial and temporal domains. Hence, a general question is how to efficiently extract the features highly correlated to deceptive behavior from both visual and audio modalities in videos and effectively fuse the information of both spatial and temporal domains. 

On the other hand, data scarcity is also a severe problem for the deception detection task. The main reason is that the objective of deception is to conceal the truth. Therefore, without some reference information from the subject, it is challenging for data collectors to determine the label information for the collected data subjectively. Until now, only one new and unique public multi-modal benchmark dataset (which contains only 121 video clips) for deception detection has been introduced in~\cite{perez2015verbal}. But, this quantity of data may still not be enough to train a deep neural network (DNN) model. Moreover, some DNN models developed with this dataset for deception detection often suffered the overfitting problem and lacked generalization ability~\cite{gogate2017deep, krishnamurthy2018deep, wu2018deception}. Hence, designing feature extraction methods with limited data becomes a critical issue.

In this paper, we propose a multi-modal framework (as shown in Fig.~\ref{fig.framework}) that extracts the emotional state-based features from visual and audio modalities to improve the method proposed in~\cite{yang2020emotion}. We futher consider the emotional states information generated from visual and audio modalities to revise their emotional transformation feature (ETF) as our new feature called emotional state transformation (EST). Our proposed EST considers the information from both spatial and temporal dimensions simultaneously. Besides, the features from the two modalities are fused to be the training data to train a series of classifiers to evaluate the performance. The experimental results demonstrate the effectiveness of the EST for multi-modal deception detection in videos. These results also show that our methods perform well even if the amount of the dataset (121 video clips) is limited. We can conclude our contributions in this work as follows: 

\begin{itemize}
    \item We propose a multimodal deception detection model framework to consider the deception detection task.
    \item We design a series of preprocessing procedures to deal with the problem in the considered dataset.
    \item We devise a voting scheme to combine the emotional state information from both visual and audio modalities to intensify the correctness of the emotional state information.
    \item Our methods can improve the overall performance of accuracy and ROC-AUC for deception detection task and have promising results compared with the state-of-the-art multi-modal deception detection methods.
\end{itemize}

The rest of this paper is organized as follows. Section~\ref{sec:related work} gives an introduction to related work. Section~\ref{sec:our proposed methods} introduces our proposed method. Experimental results are provided in Section~\ref{sec:experimental_result}, and Section~\ref{sec:conclusion} concludes this paper.

\section{Related work}
\label{sec:related work}
\subsection{Physiological characteristics based methods}
Polygraph is the most representative example of the physiological characteristics-based methods. The Polygraph test aims to measure physiological characteristics such as heart rate, respiration rate, blood pressure, and skin conductivity when the subject is asked a series of well-designed questions. Then, the measured characteristics are used as evidence to determine whether the subject is deceitful or not. But the results are usually not credible because people's judgments are often biased due to individual differences~\cite{Porter1999Vrij, gannon2009risk}. Another classic method, thermal imaging, is used to record the thermal patterns~\cite{pavlidis2002seeing}. Then, the patterns represented the distribution of blood flow of the subject body can be evidence for distinguishing the human deceits~\cite{buddharaju2005automatic}. However, although this method can output an acceptable result, the technique can not be applicable worldwide because the cost of professional devices is very expensive. Recently, the fMRI method has been applied to scan the subject's brain structure. Then, the brain images are used to find the regions related to human deception. But, again, although the detected results can reach an acceptable accuracy, some challenges such as mechanism design, experimental setting, and reliability still need to be considered~\cite{farah2014functional, langleben2013using}. To sum up, due to the testing procedures of these three physiological characteristics-based methods being public, people can easily trick the detection procedures through expert training~\cite{ganis2011lying}. 

\subsection{Computer vision based methods}
Many computer vision-based deception detection methods have been designed by considering verbal clues and non-verbal clues. For non-verbal cases, the Blob analysis is to track the head movements, hand trajectories, and body gestures~\cite{lu2005blob} as the information to classify human behavior into three distinct states. But the small dataset made the blob classifiers prone to overfitting and not applicable to new data. Recently, Ekman's research showed that people reveal some unavoidable facial clues when they are preparing deceit~\cite{ekman2009telling}. These facial clues have served as the training features for computer vision-based methods~\cite{zhang2007real, ekman1980facial}. On the other hand, for verbal cases, some researchers showed that acoustics-related features such as speaking rate, energy, pitch, range, and the identification of salient topics are meaningful with human deception~\cite{howard2011acoustic, vrij2010pitfalls}. These features, as mentioned above, can be viewed as clues to determine whether the speech is deceitful or not. In addition, linguistic-related features such as the number of words, positive and negative words, and self-reference are also associated with deceptive content~\cite{qin2005modality}.

However, the deceptive clues are not easy to detect since deception's characteristics are very subtle and varied across different subjects. Hence, the methods for detecting subtle hints such as micro facial expressions are very critical to solving this challenging problem. Moreover, features from both verbal clues and non-verbal clues are meaningful to deception detection. Therefore, combining the features from different modalities is also essential. Based on the above questions, we develop methods to efficiently and effectively extract the features from both visual and audio modalities.

\section{Our Proposed Methods}
\label{sec:our proposed methods}
To improve the performance of the deception detection task, we propose a multi-modal framework to generate the features from both visual and audio modalities. Then, we create our proposed novel feature EST based on the feature proposed in~\cite{yang2020emotion} by considering the information from both modalities. We separate the framework into three parts: feature extraction from visual modality, feature extraction from audio modality, and EST feature construction. The framework is shown in Fig.~\ref{fig.framework}.

\subsection{Feature extraction for visual modality}
The results~\cite{abouelenien2014deception} showed that facial expression is one of the most important clues to deception detection. We get inspiration from the emotional state-based feature for this task. The proposed EST feature represents the transformation between seven universal emotional states (e.g., angry, fearful, happy, sad, angry, surprised, disgusted, and neutral). However, the ground truth video clips (frame rate is 30 fps) provided in~\cite{perez2015deception} contain some unusable frames for many factors, including the face is too small to be recognized as the facial expression, the face being covered by text, file corruption, etc. Hence, we must design a series of data preprocessing methods for the aforementioned problems. For example, the Dual Shot Face Detector (DSFD)~\cite{li2018dsfd} and methods in OpenCV API~\cite{opencv_library} are used to preprocess the video frames, and some well-performed emotion recognition DNN-based methods~\cite{barsoum2016training, Meng2019FrameAN, georgescu2019local} are applied to correctly determine the emotional states for each frame in the video clips. Here, we redefine the label information of each frame with seven emotional states. Then, we use the redefined data to fine-tune the models proposed in~\cite{barsoum2016training, georgescu2019local}. As for the method in~\cite{Meng2019FrameAN}, we apply a similar way mentioned in their paper to construct the feature vector, and we use it to retrain the model to obtain the emotional state information. Then, we compare the results from the above three methods to determine the visual emotional state information. Finally, the emotional state information is incorporated with the information from the audio emotional state, which will be introduced in the next part to determine the final version of the emotional state. 

Here, we also take the features (e.g., facial displays and hand gestures) published in~\cite{perez2015deception} into account. Those features were transcribed via crowdsourcing using Amazon Mechanical Turk. The facial display feature contains the behavior information about the general facial expressions, eyebrows, eyes, mouth openness, mouth lips, and head movements. The hand gestures feature includes hand motion-based information such as hand movements and trajectories. The total dimensions of these two visual features (facial displays + hand gestures) are eighty-eight.

% -----Emotion Distribution Figure---------
\begin{figure*}[htbp]
    \centering
    \includegraphics[width=1\textwidth]{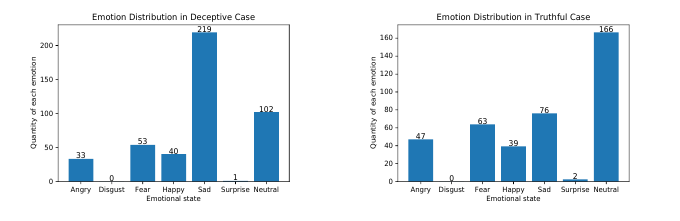}
    \caption{The emotion distribution of deceptive and truthful cases. The most frequently appearing emotional states in the deceptive case and the truthful case are sadness and Neutral, respectively.}
    \label{fig.emo1}
\end{figure*}

\subsection{Feature extraction for audio modality}
Information from audio modality is also vital for the deception detection task. Therefore, we use similar concepts and procedures for the visual modality to deal with the audio data. First, we limit the audio segments to a fixed length (e.g., 0.5-second) to satisfy the form of input of algorithms. We also redefine the label information of each audio segment with the seven emotional states. Then, we refer to the audio emotion recognition method mentioned in~\cite{kerkeni2019automatic} to determine the emotional states of each audio segment. Here, we extract the Mel-frequency cepstrum coefficients (MFCC) and modulation spectral (MS) features from the redefined data. Then, we use these features to train the support vector machine (SVM) classifier to obtain the audio emotional state information. Finally, it is considered as the reference to determine the final EST feature with the visual emotional state. The details will be introduced in the next section. 

In this part, we also use an open source software called openSMILE~\cite{inproceedings} to extract some meaningful features in audio modality. First, since the audio segments contain background noise, we use the audio processing tool~\cite{article2} to remove it. Then, we use the IS13-ComParE configuration to extract the well-performed audio features (e.g., MFCC, low-level descriptor (LLD), etc.), which are of dimension 6373 for each audio segment. In addition, we try a series of filter-based feature selection methods (e.g., Pearson's correlation, LDA, ANOVA, Chi-Square, etc.) and finally use the Pearson's correlation to reduce the feature dimension.  

\subsection{EST feature construction} 
After the emotional states from both modalities have been collected, we can determine the final version of the emotional states now. The EST feature will be affected by the results of emotion recognition during the feature extraction process. More specifically, in emotion recognition, some cases of facial expression are prone to be misclassified between two emotions (e.g., sadness or fear). This issue results in the staggered appearance of two emotions in the sequence of emotion recognition results. Suppose one emotion in the sequence of emotions in which the same emotion appears consecutively is mistaken for other emotions. In that case, it will result in two additional incorrect records in the emotional transition. Since we limited audio segments to a fixed length, the emotional state for each audio segment represented the information of fifteen frames. Therefore, we expand the emotional state fifteen times for each audio piece to align the information of visual emotional state. Then, we design a voting method to take the emotional state information from both visual and audio modalities into account to determine the final emotional state information. The idea of utilizing the majority vote is to realize the cross-comparison between the emotional state information of both modalities. In this way, we can improve the correctness of the final emotional state information. Then, our self-defined algorithms are applied to construct the EST feature with it.

Finally, we conduct a series of experiments to show the effectiveness of our proposed methods and features in the next section. The experiments contain both uni-modal and multi-modal to show the importance of both features from visual modality and audio modality.

% -----Confusion Matrix Figure---------
\begin{figure*}[t]
    \centering
    \includegraphics[width=1\textwidth]{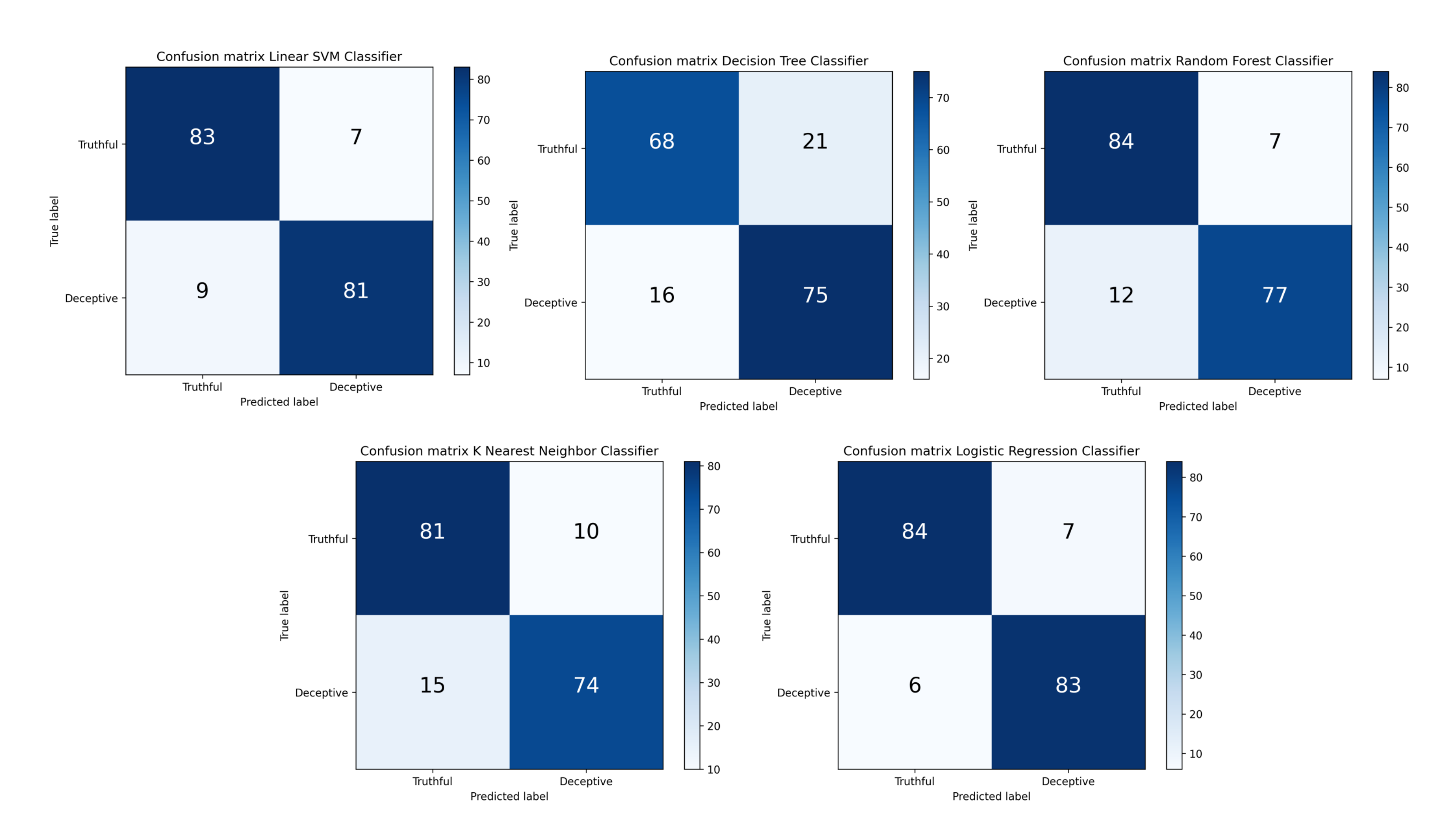}
    \caption{The confusion matrices of classifiers, including Linear Kernel Support Vector Machine (L-SVM), Decision Tree (DT), Random Forest (RF), k-Nearest Neighbor (kNN), and Logistic Regression (LR) with the multi-modal features.}
    \label{fig.cm}
\end{figure*}

%-----EST Distribution Table---------
\begin{table*}
\centering
\begin{tabular}{|c|c|c|c|c|l} 
\cline{1-2}\cline{4-5}
\textbf{EST(Top 5)}   & \textbf{Deceptive} &  & \textbf{EST(Top 5)}   & \textbf{Truthful} &   \\ 
\cline{1-2}\cline{4-5}
Sad $\to$ Fear        & 0.43               &  & Fear $\to$ Angry      & 0.36              &   \\ 
\cline{1-2}\cline{4-5}
Neutral $\to$ Sad     & 0.18               &  & Neutral $\to$ Neutral & 0.15              &   \\ 
\cline{1-2}\cline{4-5}
Happy $\to$ Neutral   & 0.10               &  & Neutral $\to$ Happy   & 0.13              &   \\ 
\cline{1-2}\cline{4-5}
Fear $\to$ Angry      & 0.08               &  & Fear $\to$ Neutral    & 0.09              &   \\ 
\cline{1-2}\cline{4-5}
Neutral $\to$ Neutral & 0.05               &  & Sad $\to$ Angry       & 0.08              &   \\
\cline{1-2}\cline{4-5}
\end{tabular}
\caption{The top 5 occurrence rates of our proposed EST.}
\label{table.est_distribution}
\end{table*}

\section{Experimental Results}
\label{sec:experimental_result}
In this section, we first describe the ground truth dataset used for our experiments and the implementation detail. Then, we perform an experimental analysis of our proposed method and compare it with several relative works.

\subsection{Dataset} 
Real-life deception detection dataset~\cite{perez2015deception} contains 121 videos, including 61 deceptive and 60 truthful trial clips. The average length of the video clips in the dataset is 28.0 seconds. Most of them in this dataset are human-edited, meaning the video clips contain transition effects and scenes irrelevant to the person we would like to observe. The unusable parts may appear at any point in the timeline of a video. We divide a video into multiple clips based on the unusable parts and label the separated clips as new clips. After we remove the unusable parts, the total number of video clips becomes ninety-four deceptive and ninety-four truthful.

\subsection{Details of our EST feature extraction} 
To extract the EST feature, we first obtain a sequence of revised emotional states determined by the emotional states detected in the visual and audio information. For detecting the visual emotional states, we extracted an image sequence from a video clip using a frame rate of 30 fps. We detect and align the faces for each frame using the DSFD algorithm proposed in~\cite{li2018dsfd}. Since multiple people may appear in the video scene simultaneously, we need to identify the target person from the crowds. We crop the image of the target person in advance as the reference and select the face most similar to the reference from the results detected by DSFD. After cropping the selected face region, we scale it to the size of 48$\times$48. Then, we obtain visual emotional states of each image with multiple facial emotion recognition model~\cite{barsoum2016training, Meng2019FrameAN, georgescu2019local}. Similarly, to detect audio EST, we split the audio file separated from the video clip into multiple 0.5-second audio segments. We obtain the emotional states of each audio segment with a speech emotion recognition algorithm~\cite{kerkeni2019automatic}. Then, we determine the revised emotional state based on the number of occurrences of emotional states in ${\bf e}_\text{v}[i]$, ${\bf e}_\text{v}[i + 1]$, and ${\bf e}_\text{ax}[i]$, where ${\bf e}_\text{v}[i]$ is the visual emotional state of frame $i$ in the corresponding video clip, and ${\bf e}_\text{ax}[i]$ is the speech emotional state corresponding to the frame. The emotional state with most occurrence is considered as the revised emotional state of frame $i$, denoted as ${\bf e}_\text{re}[i]$. If the emotional states are all different, we set the visual emotional state as the revised emotional state since the results of the visual emotional state is more reliable in our case. The processing details to obtain the revised emotional state are organized in Algorithm~\ref{alg:voting}. The construction of EST feature is based on the transition between two adjacent revised emotional states. Under the premise that the emotional state was classified into seven types in the previous stage, we classify the possible changes between two adjacent emotions into forty-nine categories based on the order of arrangement and the categories of the two emotional states. We obtain the EST feature by counting the frequency of all emotion changes in the sequence of the revised emotional states. Unlike the ETF proposed in~\cite{yang2020emotion}, if the same emotion occurs continuously, we only count it as one emotional state transition. The processing details are presented in Algorithm~\ref{alg:Framwork}.

\subsection{Analysis of emotion on deception detection}
In order to know the impact of emotion states on deception detection, for both deceptive and truthful cases, we evaluate the emotion distribution and list the five EST cases with the largest occurrence probabilities of each clip, which are shown in Fig.~\ref{fig.emo1} and Table~\ref{table.est_distribution}, respectively. We discover that emotions are mainly distributed in five emotional states, and the difference between truthful and deceptive clips appears in sadness state and neutral state. Besides, people tend to accompany a sad emotional state when they are lying. In the process of human deception, the sad to fear EST feature occurs most frequently. We think that the reason for this phenomenon is that people may undergo a heavy emotional load while deceiving and resulting in a depressive expression, or they might just pretend to be sadness as if they feel wronged.

\begin{algorithm}[htbp]  
\caption{ Framework of extracting revised emotional state feature.} 
\label{alg:voting} 
\begin{algorithmic}[1]
    \REQUIRE Video clip\\ 
    \ENSURE Revised emotional state ${\bf e}_\text{re}$\\ 
    \STATE Convert video clip to a sequence of $n$ frames;
    \STATE Crop face region of each frame and resize it into an image of size 48$\times$48; 
    \STATE Detect emotional state ${\bf e}_\text{v}[i]$ of the $i$-th image, $i = 1 \dots n$;
    \STATE Extract soundtrack from video clip;
    \STATE Split soundtrack into $m$ 0.5-second audio segment;
    \STATE Detect emotional state ${\bf e}_\text{a}[j]$ of the $j$-th audio file;
    \STATE Expand ${\bf e}_\text{a}$ by 15 times to obtain ${\bf e}_\text{ax} = {\bf e}_\text{a} \otimes {\bf 1}_{1 \times 15}$;
   
    \FOR{$i \leftarrow 1$ {\bf to} $n - 1$}
    \IF{${\bf e}_\text{v}[i + 1] = {\bf e}_\text{ax}[i]$}
    \STATE ${\bf e}_\text{re}[i] \leftarrow {\bf e}_\text{v}[i + 1]$
    \ELSE
    \STATE ${\bf e}_\text{re}[i] \leftarrow {\bf e}_\text{v}[i]$
    \ENDIF
    \ENDFOR
    \STATE ${\bf e}_\text{re}[n] \leftarrow {\bf e}_\text{v}[n]$
    \RETURN ${\bf e}_\text{re}$; 
\end{algorithmic}
\end{algorithm}

\begin{algorithm}[htbp] 
\caption{ Construction of  emotional state transformation feature.}
\label{alg:Framwork} 
\begin{algorithmic}[1] 
    \REQUIRE Revised emotional state ${\bf e}_\text{re}$\\ 
    \ENSURE Emotional state transformation feature ${\bf f}_\text{est}$\\ 
    \STATE Let $t \leftarrow 0$, emotional state transformation ${\bf T} \leftarrow {\bf 0}_{7 \times 7}$;
    \FOR{$k \leftarrow 1$ {\bf to} $n - 1$}
    \IF{${\bf e}_\text{re}[k]$ $\ne$ ${\bf e}_\text{re}[k + 1]$ or $k = 1$} 
    \STATE ${\bf T}\{{\bf e}_\text{re}[k], {\bf e}_\text{re}[k + 1]\} \leftarrow {\bf T}\{{\bf e}_\text{re}[k], {\bf e}_\text{re}[k + 1]\} + 1$
    \STATE $t \leftarrow t + 1$ 
    \ELSE
    \IF {${\bf e}_\text{re}[k]$ = ${\bf e}_\text{re}[k + 1]$ and ${\bf e}_\text{re}[k]$ $\neq$ ${\bf e}_\text{re}[k - 1]$}
    \STATE ${\bf T}\{{\bf e}_\text{re}[k], {\bf e}_\text{re}[k + 1]\} \leftarrow {\bf T}\{{\bf e}_\text{re}[k], {\bf e}_\text{re}[k + 1]\} + 1$
    \STATE $t \leftarrow t + 1$ 
    \ENDIF
    \ENDIF 
    \ENDFOR
    \STATE ${\bf f}_\text{est} \leftarrow {\bf vec}({\bf T}^{T})^{T}/t$;
    \RETURN ${\bf f}_\text{est}$; 
\end{algorithmic}
\end{algorithm}

\subsection{Evaluation setting and metrics} 
The dataset contains only fifty-eight identities, which is less than the number of video clips. Since the same identity appears in both deceptive and truthful clips, a deception detection method may suffer from overfitting to identities. To avoid the overfitting issue, we conduct experiments across different feature sets and classifiers using $K$-fold cross-validation. To test the reliability and robustness of the proposed features, we built various deception classifiers using several widely used binary classifiers, which are linear kernel support vector machine (L-SVM),  decision tree (DT), random forest (RF), k-nearest neighbor (kNN), and logistic regression (LR). After we try several cases of hyperparameter for each classifier, we obtain the following optimal setting. The regularization parameter of L-SVM is set as 1. The number of nearest neighbors used in kNN is 3. The maximum depth of the DT is 10 layers. The number of estimators used in RF is 100. For the $K$-fold cross-validation, we choose $K = 10$.

\begin{table*}
\renewcommand\arraystretch{1}
\centering
\scalebox{1}{
\begin{tabular}{|c|c|c|c|c|c|c|}
\hline
\textbf{Papers}                & \textbf{Features}       & \textbf{L-SVM}           & \textbf{DT}              & \textbf{RF}              & \textbf{kNN}             & \textbf{LR}              \\ \hline
\multirow{2}{*}{\cite{yang2020emotion}} & ETF            & 0.6575          & 0.6868          & 0.7109          & 0.6391          & 0.6296          \\ \cline{2-7} 
                      & ETF+ME         & 0.8717          & 0.7638          & 0.8099          & 0.8348          & 0.8744          \\ \hline
\multirow{2}{*}{\cite{wu2018deception}}  & IDT+MR+TR+MFCC & 0.8773          & 0.7777          & 0.8477          & -               & 0.7894          \\ \cline{2-7} 
                      & IDT+ME+TR+MFCC & 0.9065          & 0.8074          & 0.8731          & -               & 0.9221 \\ \hline
\multirow{5}{*}{Ours} & EST            & 0.8611          & \textbf{0.8444} & 0.8389          & 0.8444          & 0.8833          \\ \cline{2-7} 
                      & EST+ME         & 0.8767          & 0.7889          & 0.8544          & \textbf{0.8822} & 0.8944          \\ \cline{2-7} 
                      & EST+IS13       & 0.9178          & 0.7633          & 0.8789          & 0.8456          & 0.9122          \\ \cline{2-7} 
                      & EST+ME+IS13    & \textbf{0.9225} & 0.7964          & \textbf{0.8993} & 0.8710          & \textbf{0.9265}          \\ \cline{2-7} 
                      & EST+ME+IS13 (avg.)    & 0.9099  & 0.7775          & 0.8906 & 0.8617          & 0.9215
                      \\ \hline
\end{tabular}
}
\caption{The performance of different features measured in ROC-AUC.}
\label{table.auc}
\end{table*}

\begin{table*}
\renewcommand\arraystretch{1}
\centering
\scalebox{1}{
\begin{tabular}{|c|c|c|c|c|c|c|}
\hline
\textbf{Papers}        & \textbf{Features} & \textbf{L-SVM}   & \textbf{DT}      & \textbf{RF}      & \textbf{kNN}     & \textbf{LR}      \\ \hline
\multirow{6}{*}{\cite{perez2015deception}} & Unigrams          & 69.49\%          & 76.27\%          & 67.79\%          & -                & -                \\ \cline{2-7} 
                        & Psycholinguistic     & 53.38\%    & 50.00\%          & 66.10\%    & -          & -       \\ \cline{2-7} 
                        & Syntactic Complexity & 52.54\%    & 62.71\%          & 53.38\%    & -          & -       \\ \cline{2-7} 
                        & Facial Displays      & 78.81\%    & 74.57\%          & 67.79\%    & -          & -       \\ \cline{2-7} 
                        & Hand Gestures        & 59.32\%    & 57.62\% & 57.62\%    & -          & -       \\ \cline{2-7} 
                        & All Features         & 77.11\%    & 69.49\%          & 73.72\%    & \textbf{-} & -       \\ \hline
\multirow{3}{*}{\cite{krishnamurthy2018deep}} & DEV-vocal            & -          & -                & -          & 74.16\%    & -       \\ \cline{2-7} 
                        & DEV-visual           & \textbf{-} & -                & \textbf{-} & 75.00\%    & -       \\ \cline{2-7} 
                        & DEV-Hybrid           & -          & -                & -          & 84.16\%    & -       \\ \hline
\multirow{2}{*}{\cite{yang2020emotion}}  & ETF               & 65.00\%          & 68.46\%          & 71.15\%          & 61.15\%          & 62.45\%          \\ \cline{2-7} 
                        & ETF+ME               & 87.59\%    & 76.56\%          & 81.66\%    & 82.49\%    & 87.77\% \\ \hline
\multirow{2}{*}{Ours}                   & EST+ME+IS13       & \textbf{91.11\%} & \textbf{80.00\%} & \textbf{89.44\%} & \textbf{86.67\%} & \textbf{92.78\%}  \\ \cline{2-7} 
    & EST+ME+IS13 (avg.)       & 90.67\% & 77.33\% & 88.78\% & 85.67\% & 91.83\%
\\ \hline
\end{tabular}%
}
\caption{The comparison of the performance measured in ACCURACY.}
\label{table.acc}
\end{table*}

\subsection{Comparison of our model with existing methods} 
In this section, we first evaluate different combinations of uni-modal and multi-modal features using the ROC-AUC metric. In Table~\ref{table.auc}, we choose the most representative cases in the uni-modal method~\cite{yang2020emotion} and the multi-modal~\cite{wu2018deception} to compare with our proposed method. Here, we denote "ETF" as the features proposed in~\cite{yang2020emotion}, "ME" as the micro expression features proposed in~\cite{perez2015deception}, "MR" as the high level micro expression, "IDT" as the improved dense trajectory, "TR" as the transcripts, and "MFCC" as the Mel-frequency Cepstral Coefficients used in the state-of-the-art work~\cite{wu2018deception}, and "IS13" as the Interspeech 2013 ComParE Challenge features. For the number of features used in our experiment, our proposed EST contains 49 features, "ME" contains 88 features, and "IS13" contains 4366 features. The total number of feature is 4503. Therefore, we perform feature selection method to shrink the feature dimension to the most relevant 450 features (i.e., one-tenth of the total features) to train the classifiers. From the results, we can know that our proposed EST feature can significantly improve the performance of the ETF feature and outperform the state-of-the-art multi-modal work~\cite{wu2018deception}. For evaluating our method with other metric, we present the results measured in metric accuracy in Table~\ref{table.acc}. Here, we compare our method with the works that also measured in metric accuracy. The first work is the baseline multi-modal method proposed in~\cite{perez2015deception}. The second one is a deep neural network-based method proposed in~\cite{krishnamurthy2018deep}. They provide three kinds of result including the "DEV-vocal" that trained on vocal features, the "DEV-visual" that trained on visual features, and the "DEV-Hybrid" that trained on both vocal and visual features. Again, we can see that our approach outperforms the baseline by a huge margin and is slightly better than the DEV framework. We list both the best result and the average of the top ten results within the one hundred trials in Table~\ref{table.auc} and Table~\ref{table.acc} for comparison since the competitors did not mention how they produce their performance results (e.g., unknown number of trials to obtain their results). We can see that our average results are still comparable with those methods. In Fig.~\ref{fig.cm}, we provide confusion matrices corresponding to different classifiers to verify that our results do not have problems with skewness in data.

\section{Conclusion}
\label{sec:conclusion}
In this paper, we propose a series of data preprocessing procedures for dirty information and unusable frames in the ground truth data. Then, our proposed multi-modal deception detection framework can be used to extract features from both visual modality and audio modality. We also introduce a novel emotional state-based feature called EST feature to analyze deception detection tasks. With our designed voting method, the emotional state information from both modalities is considered to construct the EST feature. Surprisingly, our EST feature can perform very well on it even though the quantity of the dataset is very limited. We think the reason is that our EST feature includes both temporal information and spatial information from the change of emotional state in frames and audio segments. Finally, critical experiments based on our proposed methods show that we outperform the state-of-the-art method. The accuracy and ROC-AUC of deception detection in the video are improved up to 92.78\% and 0.9265, respectively. In our future work, we will take the information from textual modality into account to analyze the relationship between textual features and human deception. Also, we will design a series of data collection mechanisms for extending the diversity of data for the deception detection tasks. 

\section*{Acknowledgements}\label{sec:acknowledgment}
	This work was supported by the Ministry of Science of Technology, Taiwan, under research project number 110-2221-E-007-084-MY3.

\bibliography{aaai23}
\end{document}